# A Novel Progressive Learning Technique for Multi-class Classification

Rajasekar Venkatesan*[1] and Meng Joo Er[2]

[1,2] School of Electrical and Electronic Engineering, Nanyang Technological University, Singapore.

Email id: RAJA0046@e.ntu.edu.sg[1] EMJER@ntu.edu.sg[2]

*Abstract*— In this paper, a progressive learning technique for multi-class classification is proposed. This newly developed learning technique is independent of the number of class constraints and it can learn new classes while still retaining the knowledge of previous classes. Whenever a new class (non-native to the knowledge learnt thus far) is encountered, the neural network structure gets remodeled automatically by facilitating new neurons and interconnections, and the parameters are calculated in such a way that it retains the knowledge learnt thus far. This technique is suitable for real-world applications where the number of classes is often unknown and online learning from real-time data is required. The consistency and the complexity of the progressive learning technique are analyzed. Several standard datasets are used to evaluate the performance of the developed technique. A comparative study shows that the developed technique is superior.

## 1. INTRODUCTION

THE study on feedforward neural network (FNN) has gained prominence since the advent of back propagation (BP) algorithm [1]. Several improved and optimized variants of the BP algorithm have then been developed and analyzed [2-7]. In the past two decades, single hidden layer feedforward neural networks (SLFNs) has gained significant importance due to its widespread applications in recognition, classification and function approximation area [8-12]. Several learning techniques have been proposed since then for effective training of the SLFN [13-14]. The learning techniques can be grouped under two basic categories; Batch learning and Sequential learning [15].

Batch learning algorithms require pre collection of training data. The collected data set is then used for training the neural network. The network parameters are calculated and updated by processing all the training data together. There are several batch learning algorithms in the literature. One of the relatively new batch learning scheme called Extreme Learning Machines (ELM) is proposed by Huang et al in 2004 [16]. The special nature of ELM is that the input weights and the hidden node biases can be chosen at random [17]. A key feature of ELM is that it maintains the universal approximation capability of SLFN [17-19]. It has gained much attention to it and several research works are further made in it due to its special nature of random input weight initialization and its unique advantage of extreme learning speed [20]. The advantages of ELM over other traditional feedforward neural network are analyzed in the literature [9,21]. Many new variants and developments are made to the ELM and significant results are achieved in the approximation, classification and regression areas [22-24]. Batch learning involves processing of the complete data set concurrently for updating the weights. This technique is limited to its applications as batch learning techniques are more time consuming and requires the complete data set prior to training. On the other hand, in online/sequential learning algorithms, the network parameters are updated as and when a new training data arrives. To overcome the shortcomings of the batch learning techniques, several sequential and/or online learning algorithms are developed [25-28].

In many cases, sequential learning algorithms are preferred over batch learning algorithms as they do not require retraining whenever a new data sample is received [8]. Online-sequential learning method that combines ELM and Recursive Least Square (RLS) algorithm is later developed and is called Online-Sequential extreme learning machine (OS-ELM) [15]. Several variants of ELM and OS-ELM were developed and proposed in the literature [8,9,17,22,23,29].

The issue with existing multi-class classification techniques such as ELM and SVM is that, once they are trained to classify a specific number of classes, learning of new classes is not possible. In order to learn new class of data it requires retraining all the classes anew again.

The existing techniques require a priori information on the number of classes that will be present in the training dataset. The information on the number of classes is required to be either specified directly or is identified by analyzing the complete training data set. Based on this parameter, the network model will be designed and only the parameters or the weights of the networks are updated depending on the sequential input data. This makes the existing techniques "*static*" with respect to the number of classes it can learn.

While the existing techniques are suited for applications with pre-known dataset, it might not be well suited for applications such as cognitive robotics or those involving real-time data where the nature of training data is unknown. For such real-world and real-time data, where the number of classes to be learnt is often unknown, the learning technique must be self-developing to meet the dynamic needs. To overcome this shortcoming, a novel learning paradigm is proposed, called the "*progressive learning*".

Progressive learning is the next stage of advancement to the online learning methods. Existing online sequential techniques only learn to classify data among a fixed set of classes which are initialized during the initialization phase of the algorithm. They fail to dynamically adapt when introduced to new class/classes on the run. The progressive learning technique is independent of the number of class constraint and it can learn several new classes on the go by retaining the knowledge of previous classes. This is achieved by modifying the network structure by itself upon encountering a new class and updating the network parameters in such a way that it learns the new class and retains the knowledge learnt thus far.

The existing online sequential learning methods do not require retraining when a "new data sample" is received. But it fails when a "new class of data" which is unknown to the existing knowledge is encountered. Progressive learning technique overcomes this shortcoming by allowing the network to learn multiple new classes' alien to existing knowledge, encountered at any point of time.

## 2. PRELIMINARIES

This section gives a brief review of the ELM and the OS-ELM techniques to provide basic background information.

### 2.1 Extreme Learning Machines

A condensed overview of the batch learning ELM technique as proposed by Huang et. al. [16] is given below.

Consider there are N training samples represented as $\{(\mathbf{x_j}, \mathbf{t_j})\}$ where j varies from 1 to N, $\mathbf{x_j}$ denotes the input data vector: $\mathbf{x_j} = [x_{j1}, x_{j2}, \ldots x_{jn}]^T \in R^n$ and $\mathbf{t_j} = [t_{j1}, t_{j2}, \ldots, t_{jm}]^T \in R^m$ denotes the target class labels. Let there be P number of hidden layer neurons in the network, the output of the standard SLFN can be given as

$$\sum_{i=1}^{P} \boldsymbol{\beta}_i g_i(\mathbf{x_j}) = \sum_{i=1}^{P} \boldsymbol{\beta}_i g(\mathbf{w_i} \cdot \mathbf{x_j} + b_i) = \boldsymbol{o}_j \qquad (1)$$

where, j = 1, 2….N, $\mathbf{w_i} = [w_{i1}, w_{i2}, \ldots w_{in}]^T$ denotes the weight vector from input nodes to ith hidden node, $\boldsymbol{\beta}_i = [\beta_{i1}, \beta_{i2}, \ldots \beta_{im}]^T$ denotes the weight vector connecting i[th] hidden node to the output nodes and $b_i$ is the hidden layer bias value.

For the standard SLFN mentioned in the equation above to perform as a classifier, the output of the network should be equal to the corresponding target class of the input data given to the classifier. Hence, for the SLFN in equation 1 to be a classifier, there exist a $\boldsymbol{\beta_i}$, $g(x)$, $\mathbf{w_i}$ and $b_i$ such that

$$\sum_{j=1}^{P} \|\boldsymbol{o_j} - \boldsymbol{t_j}\| = 0 \tag{2}$$

Therefore, the equation for the output of the network can be written as,

$$\sum_{i=1}^{P} \boldsymbol{\beta_i} g(\boldsymbol{w_i} \cdot \boldsymbol{x_j} + b_i) = \boldsymbol{t_j} \tag{3}$$

where j = 1,2,…N, and $\mathbf{t_j}$ denotes the target class corresponding to the input data vector $\mathbf{x_j}$. This equation can be written in compact form as

$$\mathbf{H\beta} = \mathbf{T} \tag{4}$$

Where

$$\boldsymbol{H}(w_1, \ldots w_P, b_1, \ldots, b_P, x_1, \ldots, x_N) = \begin{bmatrix} g(w_1 \cdot x_1 + b_1) & \cdots & g(w_P \cdot x_1 + b_P) \\ \vdots & \ddots & \vdots \\ g(w_1 \cdot x_N + b_1) & \cdots & g(w_P \cdot x_N + b_P) \end{bmatrix}_{NXP} \tag{5}$$

$$\boldsymbol{\beta} = \begin{bmatrix} \beta_1^T \\ \vdots \\ \beta_P^T \end{bmatrix}_{PXm} \tag{6}$$

$$\boldsymbol{T} = \begin{bmatrix} t_1^T \\ \vdots \\ t_N^T \end{bmatrix}_{NXm} \tag{7}$$

**H** is called the hidden layer output matrix of the neural network where each column of **H** gives corresponding output of the hidden layers for a given input $\mathbf{x_i}$. The mathematical framework and the training process are extensively described in the literature [9]. The key results are restated.

*Lemma 1: [9] Given a standard SLFN with N hidden nodes and activation function $g: R \rightarrow R$ which is infinitely differentiable in any interval, for N arbitrary distinct samples $(\boldsymbol{x_i}, \boldsymbol{t_i})$, where $\boldsymbol{x_i} \epsilon R^n$ and $\boldsymbol{t_i} \epsilon R^m$, for any $\boldsymbol{w_i}$ and $b_i$ randomly chosen from any intervals of $R^n$ and $R$, respectively, according to any continuous probability distribution, then with probability one, the hidden layer output matrix $\boldsymbol{H}$ of the SLFN is invertible and $\|\boldsymbol{H\beta} - \boldsymbol{T}\| = 0$.*

*Lemma 2: [9] Given any small positive value $\varepsilon > 0$ and activation function $g: R \rightarrow R$ which is infinitely differentiable in any interval, there exists $P \leq N$ such that for N arbitrary distinct samples $(\boldsymbol{x_i}, \boldsymbol{t_i})$, where $\boldsymbol{x_i} \epsilon R^n$ and $\boldsymbol{t_i} \epsilon R^m$, for any $\boldsymbol{w_i}$ and $b_i$ randomly chosen from any intervals of $R^n$ and $R$, respectively, according to any continuous probability distribution, then with probability one, $\|\boldsymbol{H_{NxP}\beta_{PXm}} - \boldsymbol{T_{NXm}}\| < \varepsilon$.*

Thus it can be seen that for an ELM, the input weights $w_i$, and the hidden layer neuron bias $b_i$ can be randomly assigned. Training of the ELM involves estimating the output weights $\boldsymbol{\beta}$ such that the relation $\mathbf{H\beta} = \mathbf{T}$ is true.

The output weight $\boldsymbol{\beta}$ for the ELM can be estimated using the Moore-Penrose generalized inverse as $\boldsymbol{\beta} = \mathbf{H^+T}$, where $\mathbf{H^+}$ is the Moore-Penrose inverse of the hidden layer output matrix $\mathbf{H}$.

The overall batch learning algorithm of the ELM for training set of form $\{(\mathbf{x_i}, \mathbf{t_i}) | \mathbf{x_i} \epsilon R^n, \mathbf{t_i} \epsilon R^m, i = 1, \ldots N\}$ with P hidden layer neurons can be summarized as,

*STEP 1:* Random assignment of input weights $w_i$ and hidden layer bias $b_i$, i = 1, …..P.

*STEP 2:* Computation of the hidden layer output matrix H.

*STEP 3:* Estimation of output weights using $\boldsymbol{\beta} = \mathbf{H}^+\mathbf{T}$ where $\mathbf{H}^+$ is the Moore-Penrose inverse of $\mathbf{H}$ and $\mathbf{T} = [t_1, \ldots t_N]^T$.

## 2.2 Online Sequential – Extreme Learning Machine

Based on the batch learning method of the ELM, sequential modification is performed and Online Sequential-ELM (OS-ELM) is proposed in literature [15]. OS-ELM operates on online data.

In the batch learning method ELM the output weight β is estimated using the formula

$\boldsymbol{\beta} = \mathbf{H}^+\mathbf{T}$, where $\mathbf{H}^+$ is the Moore-Penrose inverse of the hidden layer output matrix $\mathbf{H}$. The $\mathbf{H}^+$ can be written as,

$$\mathbf{H}^+ = (\mathbf{H}^T\mathbf{H})^{-1}\mathbf{H}^T \tag{8}$$

As stated in [15], this solution gives the least square solution to $\mathbf{H}\boldsymbol{\beta} = \mathbf{T}$. The OS-ELM uses RLS algorithm to update the output weight matrix sequentially as the data arrives online. It has been well studied in the literature and the summary is given below.

Let $N_0$ be the number of samples in the initial block of data that is provided to the network.

Calculate $\mathbf{M_0} = (\mathbf{H_0}^T\mathbf{H_0})^{-1}$ and $\boldsymbol{\beta_0} = \mathbf{M_0}\mathbf{H_0}^T\mathbf{T_0}$.

For each of the subsequent sequentially arriving data, the output weights can be updated as

$$M_{k+1} = M_k - \frac{M_k h_{k+1} h_{k+1}^T M_k}{1 + h_{k+1}^T M_k h_{k+1}} \tag{9}$$

$$\beta_{k+1} = \beta_k + M_{k+1} h_{k+1}(t_{k+1}^T - h_{k+1}^T \beta_k) \tag{10}$$

Where k = 0,1,2…. N-$N_0$-1.

The steps in the Online-Sequential ELM based on the RLS algorithm are summarized below.

*INITIALIZATION PHASE*

*STEP 1:* The input weights and the hidden layer bias are assigned in random.

*STEP 2:* For the initial block of $N_0$ samples of data, the hidden layer output matrix $H_0$ is calculated.

$$\mathbf{H_0} = [\mathbf{h_1}, \ldots.\mathbf{h_P}]^T, \text{ where } \mathbf{h_i} = [g(w_1.\mathbf{x_i}+b_1), \ldots.g(w_P.\mathbf{x_i}+b_P)]^T, \quad i = 1,2\ldots N_0 \tag{11}$$

*STEP 3:* From the value of $H_0$, the initial values of $M_0$ and $\beta_0$ are estimated as

$$\mathbf{M_0} = (\mathbf{H_0}^T\mathbf{H_0})^{-1} \tag{12}$$

$$\boldsymbol{\beta_0} = \mathbf{M_0}\mathbf{H_0}^T\mathbf{T_0} \tag{13}$$

*SEQUENTIAL LEARNING PHASE*

*STEP 4:* For each of the subsequent sequentially arriving data, the hidden layer output vector $h_{k+1}$ is calculated.

*STEP 5:* The output weight is updated based on the RLS algorithm as,

$$M_{k+1} = M_k - M_k h_{k+1}^T (I + h_{k+1} M_k h_{k+1}^T)^{-1} h_{k+1} M_k \tag{14}$$

$$\beta_{k+1} = \beta_k + M_{k+1} h_{k+1}^T (t_{k+1} - h_{k+1}\beta_k) \tag{15}$$

The theory and the formulation behind the operation of the OS-ELM and ELM have been discussed in detail in several papers [8,9,13,15,30]. The standard variants of activation function used in ELM [31] and other special mapping functions and their variants are discussed in detail in the literature [32-35]. The other variants of ELM includes ELM Kernel [36], ELM for imbalanced data [37], ELM for noisy data [38], Incremental ELM [19], ELM ensemble [39-41] and many other variants are summarized in [31]

## 3. PROGRESSIVE LEARNING TECHNIQUE

### 3.1 Learning like children

The proposed progressive learning algorithm is adapted from the natural learning process exhibited by the children. Peter Jarvis in his book [42] has described in detail the nature of the human learning process. As opposed to traditional machine learning algorithm's training-testing cycle, human learning is a continuous process. The learning / training phase is never ending. Whenever human brain is stumbled upon with a new phenomenon, the learning resumes [42]. The key feature of human learning is that, the learning of new phenomenon does not affect the knowledge learnt. The new knowledge is leant and is added along with existing knowledge.

Though there are several online and sequential learning methods, the information of number of classes is fixed during initialization. This restricts the possibility of learning newer classes on the run. Existing machine learning algorithms fails to resume learning when an entirely new class / classes of data are encountered after the initialization. For applications such as cognitive robotics, real-world learning, etc. the system should be robust and dynamic to learn new classes on the run. The number of classes it encounters is not known beforehand. The system should be able to redesign itself and adapt to meet the learning of the new class as it arrives.

The proposed learning method introduces a novel technique of progressive learning which showcases continuous learning. The progressive learning technique enables to learn new classes dynamically on the run. Whenever a new class is encountered, the neural network "grows" and redesign its interconnections and weights so as to incorporate the learning of the new classification. Another key feature of the proposed method is that the newer classes are learnt in addition to the existing knowledge as if they were present from the beginning.

### 3.2 Proposed Algorithm

As foreshadowed, the key objective of the progressive learning technique is that it can dynamically learn new classes on the run. Suppose the network is initially trained to classify 'm' number of classes. Consider the network encounters 'c' number of new classes which are alien to the previously learnt class, the Progressive Learning Technique (PLT) will adapt automatically and starts to learn the new class by maintaining the knowledge of previously learnt classes.

The introduction of new class(es) to the network, results in changes in the dimension of the output vector and the output weight matrix. Also the newly formed matrices with increased dimension should be evaluated in such a way that it still retains the knowledge learnt thus far and also facilitates the learning of the newly introduced class(es). The method of increasing the dimension of the matrix, the weight update and matrix recalibration methods of the proposed algorithm are significantly different from the class-incremental extreme learning machine [43]. The proposed algorithm can not only learn sequential introduction of single new class, but also simultaneous (multiple new classes in same block of the online data) and sequential introduction of multiple new classes. The proposed algorithm is also independent of the time of introduction of the new class(es).

Consider there are P hidden layer neurons, and the training data is of the form $(x_i, t_i)$, the steps of the PLT algorithm are:

*INITIALIZATION PHASE*

*STEP 1:* The input weights and the hidden layer bias are assigned at random.

*STEP 2:* For the initial block of $N_0$ samples of data, the hidden layer output matrix $\mathbf{H_0}$ is calculated.

$$\mathbf{H_0} = [\mathbf{h_1}, \dots \mathbf{h_P}]^T, \quad \text{where} \quad \mathbf{h_i} = [g(w_1.\mathbf{x_i}+b_1), \dots g(w_P.\mathbf{x_i}+b_P)]^T, \qquad i = 1,2 \dots N_0 \tag{16}$$

*STEP 3:* From the value of $H_0$, the initial values of $M_0$ and $\beta_0$ are estimated as

$$\mathbf{M_0} = (\mathbf{H_0^T H_0})^{-1} \tag{17}$$

$$\boldsymbol{\beta_0} = \mathbf{M_0 H_0^T T_0} \tag{18}$$

*SEQUENTIAL LEARNING PHASE*

The subsequent data that arrives to the network can be trained either on one-by-one or chunk-by-chunk basis. Let 'b' be the chunk size. Unity value for b results in training the network on one-by-one basis.

When a new data sample/chunk of data is arrived, it can fall into either of the two categories.

i)    Absence of new class of data

ii)   Presence of new class / classes of data

If there are no new classes in the current set of data, the PLT is similar to OS-ELM and the usual process of calculating and updating the output weights is performed. The subsequent algorithm steps for the case of no new classes in current chunk of data are as follows.

*STEP 4:* The hidden layer output vector $h_{k+1}$ is calculated.

*STEP 5:* The output weight is updated based on the RLS algorithm as,

$$\boldsymbol{M_{k+1}} = \boldsymbol{M_k} - \boldsymbol{M_k} \boldsymbol{h_{k+1}^T} \left( \boldsymbol{I} + \boldsymbol{h_{k+1}} \boldsymbol{M_k} \boldsymbol{h_{k+1}^T} \right)^{-1} \boldsymbol{h_{k+1}} \boldsymbol{M_k} \tag{19}$$

$$\boldsymbol{\beta_{k+1}} = \boldsymbol{\beta_k} + \boldsymbol{M_{k+1}} \boldsymbol{h_{k+1}^T} \left( \boldsymbol{t_{k+1}} - \boldsymbol{h_{k+1}} \boldsymbol{\beta_k} \right) \tag{20}$$

If there is a new class(es) in the chunk of data arrived, a novel progressive learning technique is used to recalibrate the network to accommodate new class by retaining old knowledge.

The algorithm maintains the classes learnt thus far in a separate set. When a new data sample/block of data arrives, the data are analyzed for the class it belongs to. If the target class of new data block is equal to or a subset of existing classes, no new classification has been encountered. When the new data block's target class set is not a subset of existing classes, it means that the system has encountered new classification and a special recalibrate routine is initiated.

In the recalibration routine, the number of new classes encountered is determined and class labels are identified. Let 'c' be the number of new classes encountered. Upon identifying the number of new classes introduced, the set containing the classes learnt thus far is updated accordingly.

The neural network is redesigned with the number of output neurons increased accordingly and the interconnections redone. The weights of the new network are determined from the current and the previous weights of the old network. The weight update is made such that the knowledge learnt by the old network is retained and the knowledge of new classes is included along with it.

Consider there are P hidden layer neurons in the network, m classes of data are currently learnt by the network and b be the chunk size of the sequential learning. The introduction of 'c' new classes at any instant k+1, will modify the dimensions of the output weight matrix $\beta$ from $\beta_{PXm}$ to $\beta_{PXm+c}$.

The output weight matrix $\beta$ is of critical importance in ELM based networks. Since the input weights and the hidden layer bias are randomly assigned, the values in the $\beta$ matrix control the number of classes learnt and the accuracy of each class. The algorithm steps are continued as follows.

*STEP 4:* The values of $\beta_{PXm+c}$ are calculated based on the current values of $\beta_{PXm}$, $(h_k)_{bXP}$ and $(M_k)_{PXP}$.

The current **β** matrix is of the dimension $(\beta_k)_{PXm}$ and 'c' new classes are introduced. Therefore, to accommodate the output weight matrix for the increased number of output layer neurons, the **β** matrix is transformed to $\widetilde{\beta_k}$ as given in equation.

$$\widetilde{\beta_k} = (\beta_k)_{PXm} \; I_{mXm+c} \tag{21}$$

Where $I_{mXm+c}$ is a rectangular identity matrix of dimension m X m+c.

$$\widetilde{\beta_k}_{PXm+c} = (\beta_k)_{PXm} \begin{bmatrix} 1 & 0 & \dots & 0 \\ 0 & 1 & \dots & 0 \\ 0 & 0 & \dots & 0 \\ 0 & 0 & \dots & 0 \end{bmatrix}_{mXm+c} \tag{22}$$

$$\widetilde{\beta_k}_{PXm+c} = [(\beta_k)_{PXm} \quad O_{PXc}]_{PXm+c} \tag{23}$$

Where $O_{PXc}$ is zero matrix.

Upon extending the weight matrix to accommodate the increased number of output neurons, the learning learnt thus far has to be incorporated in the newly upgraded weight matrix. Appending zero matrix is a trivial way to increase the dimensions. The matrix values have to be updated such that the network retains the knowledge of existing classes and can learn new classes as if they were available from the beginning of the training phase.

From equation 18, it can be seen that, the error difference between the target class $t_{k+1}$ and the predicted class $h_{k+1}\beta_k$ is scaled by a learning factor and is added to $\beta_k$. Since 'c' new classes are introduced only at the k+1th time instant, for the initial k data samples, the target class label value corresponding to the new class is -1. Therefore, the k-learning step update for the 'c' new classes $((\Delta\beta_k)_{PXc})$ can be written as,

$$(\Delta\beta_k)_{PXc} = (M_k)_{PXP} \; (h_k^T)_{PXb} \begin{bmatrix} -1 & \cdots & -1 \\ \vdots & \ddots & \vdots \\ -1 & \cdots & -1 \end{bmatrix}_{bXc} \tag{24}$$

$$(\Delta\beta_k)_{PXc} = -(M_k)_{PXP} \; (h_k^T)_{PXb} \; J_{bXc} \tag{25}$$

where $J_{bXc}$ is an all-ones matrix.

$$J_{bXc} = \begin{bmatrix} 1 & \cdots & 1 \\ \vdots & \ddots & \vdots \\ 1 & \cdots & 1 \end{bmatrix}_{bXc} \tag{26}$$

The k-learning step update for the new classes is then incorporated with the $\widetilde{\beta_k}_{mXm+c}$ to provide the upgraded $(\beta_k)_{PX(m+c)}$ matrix which is recalibrated to adapt learning 'c' new classes.

$$(\Delta\widetilde{\beta_k})_{PXm+c} = [O_{PXm} \quad -(M_k)_{PXP} \; (h_k^T)_{PXb} \; J_{bXc}] \tag{27}$$

The recalibrated output weight matrix $(\beta_k)_{N'X(N'+c)}$ is calculated as,

$$(\beta_k)_{PX(m+c)} = \widetilde{\beta_k}_{PXm+c} + (\Delta\widetilde{\beta_k})_{PXm+c} \tag{28}$$

Upon simplification, $(\beta_k)_{PX(m+c)}$ can be expressed as,

$$(\beta_k)_{PX(m+c)} = [ (\beta_k)_{PXm} \quad (\Delta\beta_k)_{PXc} ] \tag{29}$$

$(\beta_k)_{PXm}$ represents the knowledge previously learnt. The dimension of β is increased from m to m+c. As opposed to populating the increased dimension with identity matrix values, the new entries $(\Delta\beta_k)_{PXc}$ are calculated in such a way that the newly introduced classes will appear to the neural network as if they are present from the beginning of the training procedure and the training data samples thus far does not belong to the newly introduced class.

The network is recalibrated such that the $(\Delta\boldsymbol{\beta}_k)_{PXc}$ matrix represents the learning of the new class from the beginning of the training phase to the current data sample considering that none of the previous data samples belong to the newly introduced class. i.e. The $(\Delta\boldsymbol{\beta}_k)_{PXc}$ is computed which is equivalent to the k-learning step equivalent of the 'c' new classes from the beginning of the training phase.

Therefore the updated $(\boldsymbol{\beta}_k)_{PX(m+c)}$ matrix represents the network with (m+c) classes with 'm' previously existing classes and 'c' new classes.

*STEP 5:* The hidden layer output vector $h_{k+1}$ is calculated.

*STEP 6:* The output weight matrix of increased dimension to facilitate learning of new class is updated based on the RLS algorithm as,

$$M_{k+1} = M_k - M_k h_{k+1}^T (I + h_{k+1} M_k h_{k+1}^T)^{-1} h_{k+1} M_k \qquad (30)$$

$$\boldsymbol{\beta}_{k+1} = \boldsymbol{\beta}_k + M_{k+1} h_{k+1}^T (t_{k+1} - h_{k+1} \boldsymbol{\beta}_k) \qquad (31)$$

Whenever a new class(es) are encountered, the training resume learning the new class/classes by retaining the existing knowledge. The algorithm also supports recalibration with multiple new classes introduced simultaneously and sequentially. Also, the new classes can be introduced at any instant of time and any number of times to the network.

The algorithm of the progressive learning technique (PLT) is summarized in Fig. 1.

## 4  EXPERIMENTATION

Proposed progressive learning algorithm exhibit "dynamic" learning of new class of data. Current multiclass classification algorithms fails to adapt when encountered with new class and hence the accuracy drops when introduced with one or more new classes. The proposed algorithm redesigns itself to adapt to new classifications and still retaining the knowledge learnt thus far.

The proposed progressive learning algorithm is tested with several real world and standard datasets. The standard datasets are in general uniformly distributed. But to test the performance of progressive learning effectively, it should be presented with conditions where new classes are introduced in a non – uniform manner at different time instants. Hence the standard datasets cannot be used directly to test the progressive learning algorithm efficiently. The datasets should be in such a way that only a subset of classes is available for training initially and new classes should be introduced at arbitrary time instances during the latter part of training. Thus, some of the standard datasets are modified and used for testing the proposed algorithm.

By default, classification problems involve two classes: 1. Presence of class and 2. Absence of class. These are the two trivial classes that are available in any of the classification problem. Since the minimum number of classes in a classification is two, learning of new classes is absent in bivariate datasets. For binary classification datasets, since there are only two classes and no new classes are introduced, the proposed algorithm performs similar to the existing online sequential algorithm. The unique feature of progressive learning is clearly evident only in multiclass classification.

Thus the proposed algorithm is tested with multiclass classification datasets such as iris, balance scale, waveform, wine, satellite image, digit and character datasets. The specifications of the datasets are shown in TABLE 1. The proposed technique is experimented with both balanced and unbalanced datasets. Balanced dataset is one in which each of the class has equal or almost equal number of training data. Unbalanced dataset is a skewed dataset where a subset of classes has a high number of training samples and other classes have fewer training samples. The number of hidden layer neurons for the experimentation is chosen such that the overfitting problem is mitigated. The test dataset consists data samples corresponding to all the class labels used for progressive learning of the network.

---

**Algorithm : Progressive Learning Technique for Multi-class Classification**

---

1.  The parameters of the network are initialized

2.  The raw input data is processed for classification

3.  ELM Training – Initial phase

    Processing of initial block of data

    $$M_0 = (H_0{}^T H_0)^{-1}$$

    $$\beta_0 = M_0 H_0{}^T Y_0$$

4.  ELM Training – Sequential phase

    Case 1: No new classes are introduced:

    $$M_{k+1} = M_k - \frac{M_k h_{k+1} h_{k+1}^T M_k}{1 + h_{k+1}^T M_k h_{k+1}}$$

    $$\beta_{k+1} = \beta_k + M_{k+1} h_{k+1} (Y_{k+1}^T - h_{k+1}^T \beta_k)$$

    Case 2: 'c' new classes are introduced:

    $$(\beta_k)_{N' X (m+c)} = [\ (\beta_k)_{N' Xm} \quad (\Delta\beta_k)_{N'Xc}\ ]$$

    $$(\Delta\beta_k)_{N'Xc} = (M_k)_{N'XN,} \ (h_k^T)_{N'Xb} \begin{bmatrix} -1 & \cdots & -1 \\ \vdots & \ddots & \vdots \\ -1 & \cdots & -1 \end{bmatrix}_{bXc}$$

    $$M_{k+1} = M_k - \frac{M_k h_{k+1} h_{k+1}^T M_k}{1 + h_{k+1}^T M_k h_{k+1}}$$

    $$\beta_{k+1} = \beta_k + M_{k+1} h_{k+1} (Y_{k+1}^T - h_{k+1}^T \beta_k)$$

5.  ELM Testing

    Estimation of raw output values using Y = Hβ

    Class corresponding to the index of maximum value of Yi is the predicted target class

---

Fig. 1. Algorithm of Progressive Learning Technique

The proposed algorithm also works for introduction of multiple new classes. The number of classes can be increased from 2 to 3, and then from 3 to 4 and 4 to 5 and so on. For testing multiple new classes, the proposed method is tested with character recognition dataset which is described in the latter part of this section. The introduction of multiple classes both sequentially and simultaneously at multiple time instances are experimented and verified.

## 5    RESULTS AND DISCUSSIONS

The functionality of the technique, consistency and complexity are the three key features to be tested for any new technique. The functional testing is used to validate that the proposed algorithm is functional and results in its expected behavior. The functionality of the technique is tested using iris, waveform and balance-scale datasets. The operational working of the concept of progressive learning in the proposed algorithm is tested in the functionality test. Consistency is another key feature that is essential for any new technique. The proposed algorithm should provide consistent results for multiple trials with minimal variance. Being an ELM based algorithm, the consistency of the proposed method across several trials of same dataset and also the consistency across 10-fold cross validation are tested. Complexity analysis is essential for a new technique. The number of operations performed and calculations involved in the proposed method is computed and is compared against the existing method. Also the performance of the proposed algorithm is evaluated by introducing new classes at different time instances (1. Very early during training, 2. In the middle of training and 3. Towards the end of training) are evaluated. Both sequential and simultaneous introduction of new classes are experimented and results are analyzed and discussed.

TABLE I

SPECIFICATIONS OF MULTICLASS CLASSIFICATION DATASETS

| Dataset | Number of classes | Number of features/attributes | Remarks |
|---------|-------------------|-------------------------------|---------|
| Iris dataset | 3 | 4 | Basic benchmark dataset |
| Balance scale dataset | 3 | 4 | Benchmark dataset for unbalanced data |
| Waveform dataset | 3 | 21 | Basic benchmark dataset |
| Wine dataset | 3 | 13 | Basic benchmark dataset |
| Satellite image dataset | 6 | 36 | Basic benchmark dataset |
| Digit dataset | 10 | 64 | Basic benchmark dataset |
| Character dataset 1 | 4 | 17 | Dataset for sequential introduction of two new classes |
| Character dataset 2 | 5 | 17 | Dataset for sequential introduction of three new classes |
| Character dataset 3 | 5 | 17 | Dataset for simultaneous introduction of new classes |

## 5.1 Functionality

The proposed technique is experimented with iris, waveform and balance scale datasets to verify the basic intended functionality of the technique. The iris dataset consists of three classes which are uniformly distributed over the 150 instances. To facilitate testing of progressive learning, the dataset is redistributed such that first 50 samples consists of only two classes (sentosa, versicolor) and the third class (virginica) is introduced only after the 51st sample. This type of redistribution closely emulates the real time scenario of encountering a new class on the run. The ability of the proposed algorithm to recognize, adapt and learn the new class can be verified by this testing. The distribution details of the dataset used is given in TABLE 2.

TABLE 2

SPECIFICATIONS OF IRIS DATASET

| Data range | Number of classes | New class added | Point of introduction of new class | Class labels |
|---|---|---|---|---|
| 1–50 | 2 | - | - | Sentosa |
| | | | | Versicolor |
| 50–150 | 3 | 1 | 51 | Sentosa |
| | | | | Versicolor |
| | | | | Virginica |

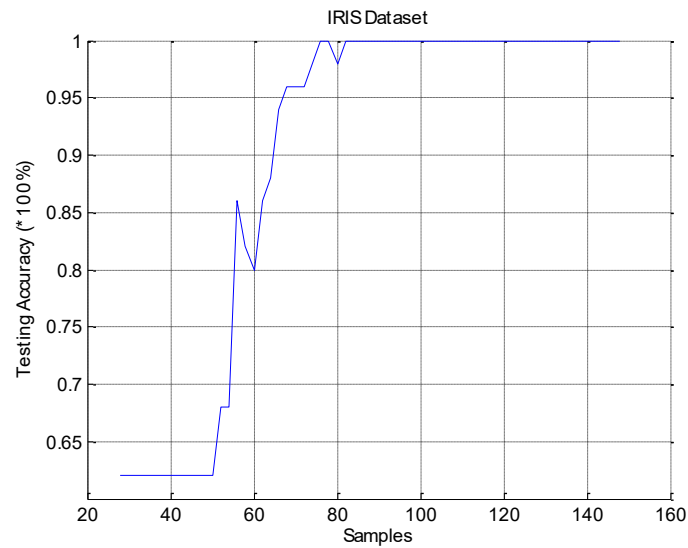

Fig. 2. Testing Accuracy in Iris Dataset

The Progressive Learning algorithm is tested with the specified iris dataset and the learning curve or the testing accuracy graph is plotted. The result obtained is shown in Fig. 2. The testing accuracy is continuously calculated with the test data set for every new training data. It can be seen from the graph that until the sample index 50, the testing accuracy is only 66.6 %. This implies that the system has learnt only two of the three classes thus far. When the third new class is introduced in the 51st sample, the system recognizes the introduction of a new class and recalibrates itself by sufficiently increasing the number of neurons in the output layer. The weight matrix is suitably increased in dimension and the weights are updated based on the special recalibration technique proposed. A new network structure and weight parameters are formed from the current network parameters and the data obtained from the new class. In the forthcoming iterations, the system then trains for recognition of the new class in addition to previously learnt classes and reaches a steady state testing accuracy. This process results in the sudden rise in the testing accuracy of the network, which then settles at a final testing accuracy value. This sudden increase in the testing accuracy is due to the fact that the network can now recognize the newly encountered class of data.

The same procedure is repeated for waveform and balance scale dataset. The dataset specifications of the waveform and balance scale dataset are shown in TABLE 3-4. The result obtained by the progressive learning method is shown in Fig. 3-4 respectively.

TABLE 3

SPECIFICATIONS OF WAVEFORM DATASET

| Data range | Number of classes | New class added | Point of introduction of new class | Class labels |
|---|---|---|---|---|
| 1–1500 | 2 | - | - | Waveform 1 |
| | | | | Waveform 2 |
| 1501–3000 | 3 | 1 | 1501 | Waveform 1 |
| | | | | Waveform 2 |
| | | | | Waveform 3 |

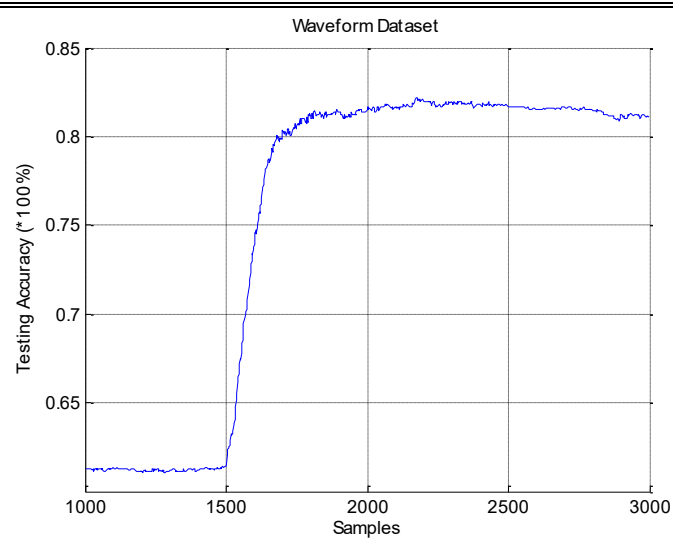

Fig. 3. Testing Accuracy of Waveform Dataset

TABLE 4

SPECIFICATIONS OF BALANCE SCALE DATASET

| Data range | Number of classes | New class added | Point of introduction of new class | Class labels |
|---|---|---|---|---|
| 1–350 | 2 | - | - | Left |
| | | | | Right |
| 351–1100 | 3 | 1 | 351 | Left |
| | | | | Right |
| | | | | Balanced |

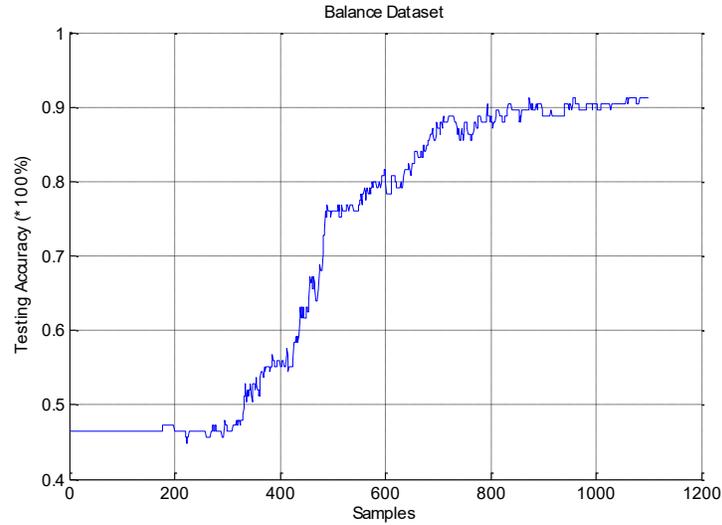

Fig. 4. Testing Accuracy of Balance dataset

From the test results, the expected behavior of the progressive learning method is verified. The result shows that the algorithm is able to learn new classes dynamically on the run and the learning of new class does not significantly affect the accuracy of the classes previously learnt. The consistency and performance of the proposed method is evaluated using six benchmark datasets.

## 5.2 Consistency

Consistency is a critical characteristic to be tested for any new technique. The proposed technique is verified for its consistency in its results. Consistency is a key virtue that any technique should exhibit. The learning technique which provides inconsistent results is not reliable for practical applications. Being an ELM based technique, the input weights and the hidden layer bias values are initialized at random. Hence multiple executions of the same dataset and same specification results in different results. Therefore, the same dataset with same specification is executed multiple times to determine the consistency across multiple executions. The consistency results of repeated multiple executions of the three datasets are shown in TABLE 5.

TABLE 5

CONSISTENCY: ACROSS MULTIPLE TRIALS (10 TRIALS)

|  | Testing Accuracy (%) |
| --- | --- |
| Iris Dataset | 99.4 ± 0.9660 |
| Waveform Dataset | 83.9 ± 1.2589 |
| Balance Scale Dataset | 91.6 ± 1.0557 |
| Wine Dataset | 97.9 ± 0.9285 |
| Satellite Image Dataset | 89.6 ± 1.1640 |
| Digit Dataset | 97.1 ± 0.7854 |

Cross validation is the most common method to evaluate the consistency of any given technique. The proposed algorithm is tested with each of the datasets for 5-fold cross validation (5-fcv) and 10-fold cross validation (10-fcv) and the resulting testing accuracy is tabulated. TABLE 6 gives the consistency of the proposed algorithm for cross validation performance. It can be seen from the table that the proposed algorithm is consistently accurate in each of the attempts. The deviation of the testing accuracy is in order of about 1 % from the mean value which is nominal. Thus, the results show that the proposed method gives consistent and reliable testing accuracy for both balanced and unbalanced datasets.

TABLE 6

CONSISTENCY: 5-FOLD CROSS VALIDATION AND 10-FOLD CROSS VALIDATION

|  | *5-fcv* | *10-fcv* |
|---|---|---|
| Iris Dataset | 99 ± 1.0954 | 99.4 ± 1.0544 |
| Waveform Dataset | 84.1 ± 1.2589 | 83.6 ± 1.3658 |
| Balance Scale Dataset | 91.8 ± 1.0557 | 91.2 ± 1.3847 |
| Wine Dataset | 97.5 ± 1.5376 | 97.9 ± 1.4625 |
| Satellite Image Dataset | 89.4 ± 1.4618 | 89.8 ± 1.5537 |
| Digit Dataset | 97.2 ± 1.0441 | 96.9 ± 1.0683 |

## 5.3 Computational Reduction

The number of computations required for the proposed progressive learning technique is analyzed and compared with the existing OS-ELM method. Though learning of new classes dynamically on the run causes overhead to the computations and seemingly increases the complexity of the technique, the actual computational complexity of the proposed technique is lesser than the OS-ELM method. The decrease in complexity is due to two reasons.

1. The overhead computations responsible for increasing the number of output neurons, creating new interconnections and recalibration of weights occur only during the samples when a new class is introduced. Thus , the recalibration routine is invoked only when there is a new class, henceforth causing minimal increase in the computation complexity. For example, when only one new class is introduced, the recalibration procedure is invoked only once.

2. The progressive learning method also provides another distinct advantage. Since the new classes are learnt dynamically, it results in lesser number of weight calculations when compared with other static online sequential training techniques like OS-ELM.

For example, in the iris dataset considered, the traditional algorithm needs to update the weight for all three output neurons for the entire 150 instances of the training set. But in the proposed method, there are only two output neurons till the occurrence of the third class. The third output neuron is introduced only during the recalibration stage triggered by the introduction of a new class. Thus, the number of weight calculations is effectively reduced. This effectively reduces the number of computations performed and thereby reducing the computational complexity. The reduction in number of weight calculations is shown in TABLE 7. The number of computations in the OS-ELM is normalized to 100 and the computational complexity of OS-ELM and the progressive learning method are compared in Fig. 5.

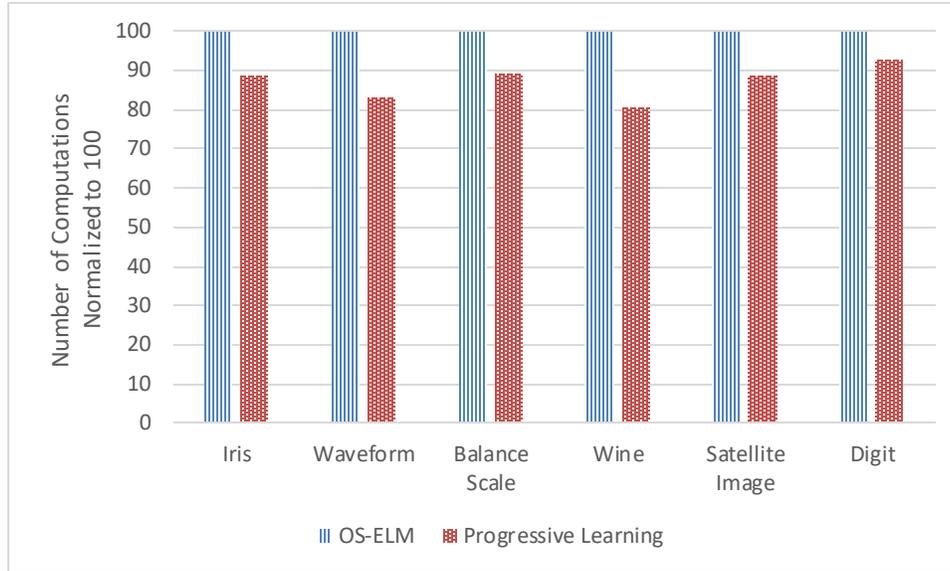

Fig. 5. Comparison of Computational Reduction

TABLE 7

REDUCTION IN NUMBER OF CALCULATIONS BY THE PROPOSED METHOD

| | No. of weight calculations in OS-ELM ( *nHidden) | Point of introduction of new class | No. of weight calculations in proposed method  (* nHidden) | % of calculations saved |
|---|---|---|---|---|
| Iris dataset | 150 * 3 | 51 | (50 * 2) + (100 * 3) | 11.11 % |
| Waveform dataset | 3000 * 3 | 1501 | (1500 * 2) + (1500 * 3) | 16.67 % |
| Balance scale dataset | 1100 * 3 | 351 | (350 * 2) + (750*3) | 10.61 % |
| Wine dataset | 120*3 | 71 | (70 * 2) + (50 * 3) | 19.44 % |
| Satellite image dataset | 4500 * 6 | 3001 | (3000 * 5) + (1500 * 6) | 11.11% |
| Digit dataset | 4000 * 10 | 3001 | (3000 * 9) + (1000 * 10) | 7.5 % |

Though the new classes are learnt only from halfway through the datasets, the testing accuracy of the algorithm is nearly maintained or even improved when compared to algorithms with a static number of classes. The reason for the change in accuracy is due to the fact that new classes are learnt on the run after the learning of previous classes. If the previously learnt classes and the new class are fairly distinctive the learning accuracy will be improved. On other hand, in some cases due to the feature set of the learnt class and new class, the learning of the new class will affect the existing knowledge but only to a little extent thereby marginally reducing the overall accuracy.

The testing accuracy of the proposed algorithm is compared with the existing OS-ELM and its variants such as voting based OS-ELM (VOS-ELM), enhanced OS-ELM (EOS-ELM), robust OS-ELM (ROS-ELM), robust bayesian ELM (RB-ELM) and generalized pruning ELM (GP-ELM) and is tabulated as shown in TABLE 8. It can be seen from the table that despite learning the new classes dynamically at a later stage of training, the testing accuracy is either improved or maintained nearly equal to the testing accuracy of the OS-ELM based methods. But the proposed method provides two key advantages over the existing methods.

1. Reduction in computational complexity.

2. Flexibility to learn new classes at any instant of time.

From the results obtained thus far, it is evident that the proposed progressive learning algorithm learns new class of data in a dynamic way.

TABLE 8

COMPARISON OF TESTING ACCURACY (%)

| | OS-ELM | VOS-ELM | EOS-ELM | ROS-ELM | RB-ELM | GP-ELM | Proposed method |
|---|---|---|---|---|---|---|---|
| *IRIS dataset* | 98 | 99.2 | 100 | 100 | 100 | 100 | 100 |
| *Waveform dataset* | 84.2 | 84.8 | 84.6 | 85.1 | 84.3 | 84.7 | 83.9 |
| *Balance scale dataset* | 90.7 | 91.1 | 90.8 | 91.4 | 90.9 | 91.3 | 91.6 |
| *Wine dataset* | 97.2 | 97.5 | 97.4 | 98.0 | 97.1 | 97.6 | 97.9 |
| *Satellite image dataset* | 88.9 | 89.2 | 89.0 | 89.1 | 89.5 | 89.8 | 89.6 |
| *Digit dataset* | 96.6 | 96.8 | 96.5 | 96.9 | 97.1 | 97.3 | 97.1 |

## 5.4 Introduction of New Class at Different Time Instants

The new class is introduced at different stages of the training period and its effect on learning rate is analyzed. In order to analyze the response, three different test cases are experimented and performance is measured. The new class is introduced at three different time instances. 1. Very early during training, 2. In the middle of training, 3. Towards the end of training. The testing accuracy curve in each of the test case is plotted and the results are evaluated and compared. The point of introduction of a new class for each of the test case is tabulated and is given in TABLE 9. The performance of the proposed network for each of the test cases is given in Fig. 6. It can be seen from the figure that, independent of the point of introduction of a new class to the system, the network is capable of learning the new class and the final steady state testing accuracy is the same across the test cases.

TABLE 9

POINT OF INTRODUCTION OF NEW CLASS

| Test cases | Point of introduction of new class (Total number of samples = 150) |
|---|---|
| Very Early | 6 |
| In the Middle | 71 |
| Towards the End | 131 |

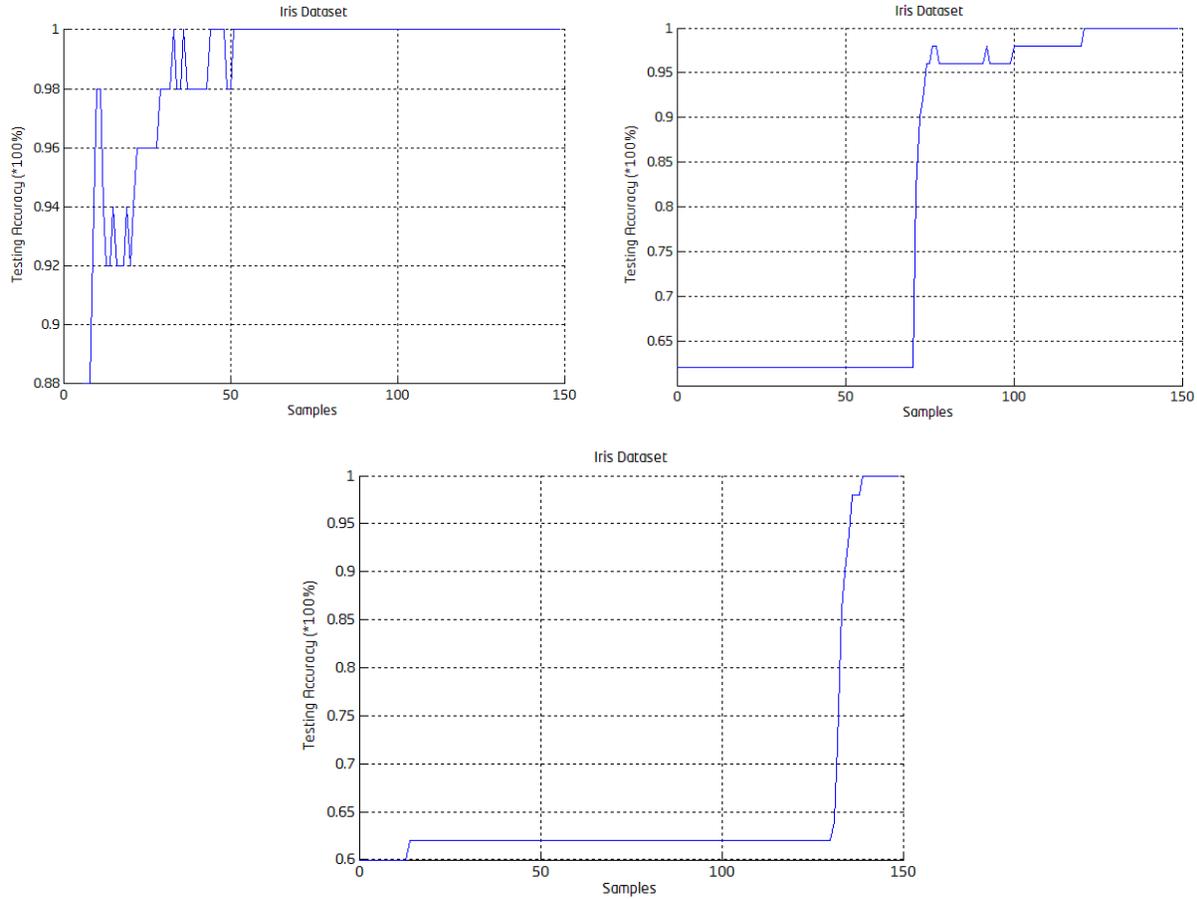

Fig. 6: Introduction of new class in training (a) Very early, (b) In the middle and (c) Towards the end

## 5.5  Multiple New Classes

The performance of the proposed technique when introduced with multiple new classes both sequentially and simultaneously is discussed in this section. Learning of multiple new classes by the proposed algorithm is tested by using the Character recognition dataset. Several combinations of tests are made such as

1. Sequential introduction of 2 new classes (4 classes)

2. Sequential introduction of 3 new classes (5 classes)

3. Simultaneous introduction of 2 new classes along with one new class sequentially (5 classes)

The performance of the proposed algorithm on each of the test case is observed.

### 5.5.1 Sequential Introduction of 2 new classes

Character dataset with 4 classes (A, B, C and D) is used to test the sequential introduction of two new classes in the proposed algorithm. The dataset is redistributed to meet the testing requirements for progressive learning. The specifications of the dataset are given in TABLE 10.

TABLE 10

SPECIFICATIONS OF CHARACTER DATASET FOR 2 NEW CLASSES

| Data range | Number of classes | New class added | Class labels |
|---|---|---|---|
| 1 – 800 | 2 | - | A and B |
| 801 – 1600 | 3 | C | A, B and C |
| 1601 – 3096 | 4 | D | A,B,C and D |

Initially the network is sequentially trained with only two classes A and B up to 800 samples. A new class 'C' is introduced to the training data in the 801st sample and a fourth class 'D' is introduced as 1601st sample. The proposed algorithm identifies both the new classes and recalibrates itself each time and continues learning. This results in two sudden rise in the learning curve of the network. The first rise corresponding occurring at 801st sample corresponds to the learning of class 'C' and the second rise occurring at 1601st sample corresponds to learning of class 'D'. The learning curve graph is shown in Fig. 7.

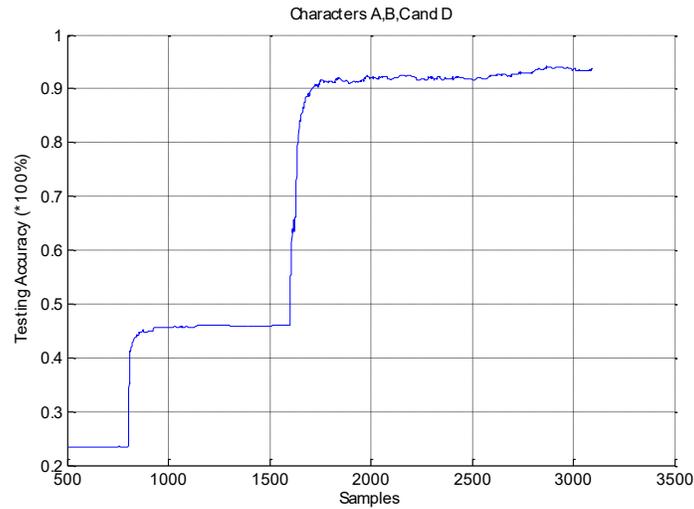

Fig. 7. Sequential learning of two new classes

### 5.5.2 Sequential introduction of 3 new classes

Character dataset with 5 classes (A, B, C, D and E) is used for testing sequential introduction of 3 new classes. The network is initially trained to recognize only two classes. Three new classes (C, D and E) are introduced one after another after the initial training of two classes. The specifications of the dataset are shown in TABLE 11.

TABLE 11

SPECIFICATIONS OF CHARACTER DATASET FOR THREE NEW CLASSES

| Data range | Number of classes | New class added | Class labels |
|---|---|---|---|
| 1 – 800 | 2 | - | A and B |
| 801 – 1600 | 3 | C | A, B and C |
| 1601 – 2000 | 4 | D | A,B,C and D |
| 2001 – 3850 | 5 | E | A,B,C,D and E |

Each of the new classes is introduced sequentially at later time instants and the algorithm adapts to new class each time and also maintains the testing accuracy at the same level. The testing accuracy curve is shown in Fig. 8.

To verify that learning of each new class is independent of previously learnt classes, the overall testing accuracy is broken down into individual testing accuracy of each of the classes and is shown in Fig. 9. It can be seen that, the testing accuracy of each of the classes remains over 90%. Also, whenever a new class is introduced, a new learning curve is formed which contributes towards the overall accuracy along with the existing classes.

The network is initially trained with two classes A and B. Third class C is introduced after 800 samples and the learning curve of the class C is shown in black line. Another new class 'D' who's testing accuracy as shown in red is introduced after the 1600 sample. A fifth class, 'E' is introduced in the 2001st sample and its learning curve is shown in light blue.

From the graph it can be seen that each class introduced is learnt anew without affecting much the existing knowledge. The learning accuracy of each of the classes is collectively responsible for the overall accuracy of the network. Further, it can be seen that the testing accuracy of each of the classes is over 90% and the overall accuracy of 94% is achieved.

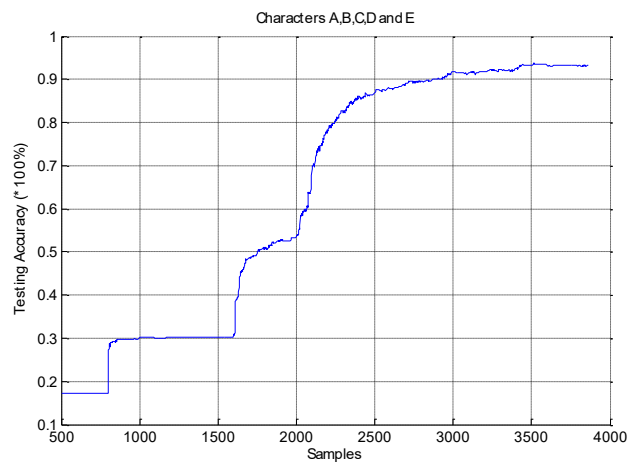

Fig. 8. Sequential learning of three new classes

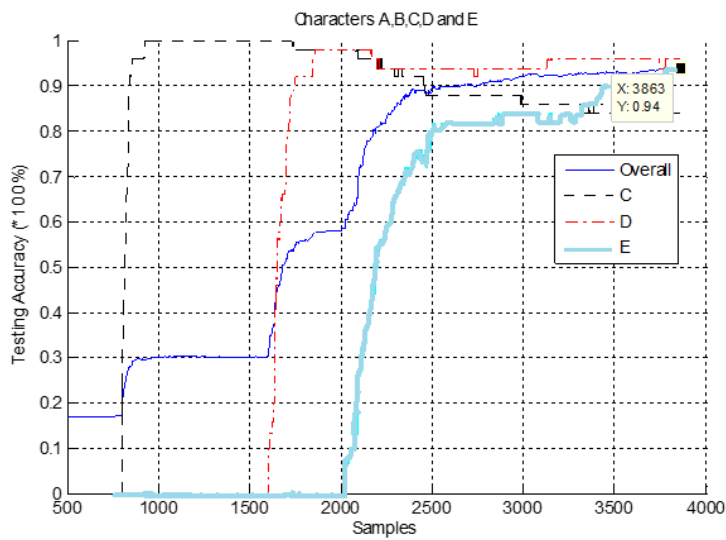

Fig. 9: Individual and Overall Testing Accuracy – Sequential Introduction

The testing accuracy obtained by introducing one, two and three new classes is summarized in TABLE 12.

From the table it can be observed that learning of multiple new classes does not affect the testing accuracy of previously learnt class. Hence this method can be used to learn a large number of multiple new classes in a progressive manner without affecting the testing accuracy of previously learnt classes.

TABLE 12

SUMMARY OF TESTING ACCURACY FOR SEQUENTIAL INTRODUCTION OF MULTIPLE NEW CLASSES

| Number of classes introduced sequentially | Testing Accuracy |
|---|---|
| Two base class + One new class | 93.8 % |
| Two base class + Two new classes | 93.7 % |
| Two base class + Three new classes | 94 |

### 5.5.3 Simultaneous introduction of new classes

To verify that the proposed algorithm performs effectively when multiple classes are introduced simultaneously (introduced in the same block), character dataset with specifications as shown in TABLE 13 is used. Here, the two classes C and D are introduced together and the new class E at a later stage. The testing accuracy is shown in Fig. 10.

TABLE 13

SPECIFICATIONS OF CHARACTER DATASET FOR SIMULTANEOUS NEW CLASSES

| Data range | Number of classes | New class added | Class labels |
|---|---|---|---|
| 1 – 800 | 2 | - | A and B |
| 801 – 2000 | 4 | C,D | A, B, C and D |
| 2001 – 3850 | 5 | E | A,B,C,D and E |

The first rise observed at the sample instant of 800 in the testing accuracy curve corresponds to the introduction of two new classes (characters C and D). The algorithm identifies both the new classes and recalibrates to facilitate multiple class addition. The second rise in the curve corresponds to the introduction of the third class (character E).

In order to show that the previous knowledge is retained and new knowledge is added along with the existing, the testing accuracy is split up for each of the five alphabets and is shown in Fig. 11.

It can be seen that, two new learning curves corresponding to each new class C and D is introduced in the 800[th] sample. Both of the newly introduced classes are learnt simultaneously along with the existing classes A and B. The learning curve at 1600[th] sample index corresponds to the introduction of class E.

Also, from the graph it is clear that the learning of additional classes does not significantly affect the testing accuracy of the classes previously learnt. Thus, enabling the proposed algorithm to learn multiple new classes both sequentially and simultaneously in a progressive manner.

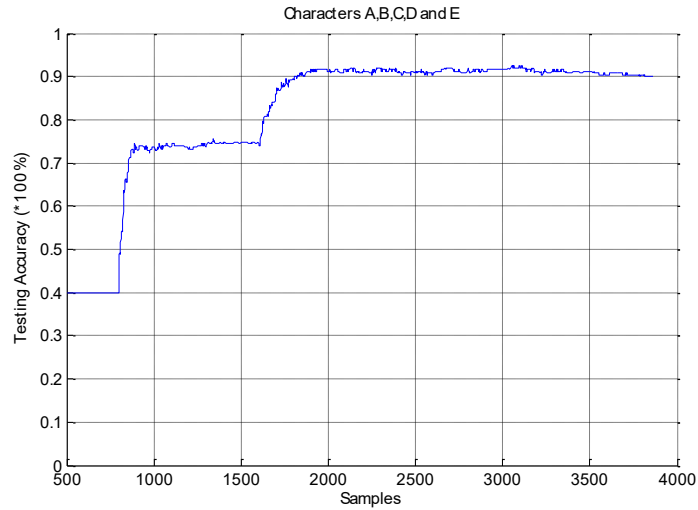

Fig. 10. Testing Accuracy for Simultaneous new classes

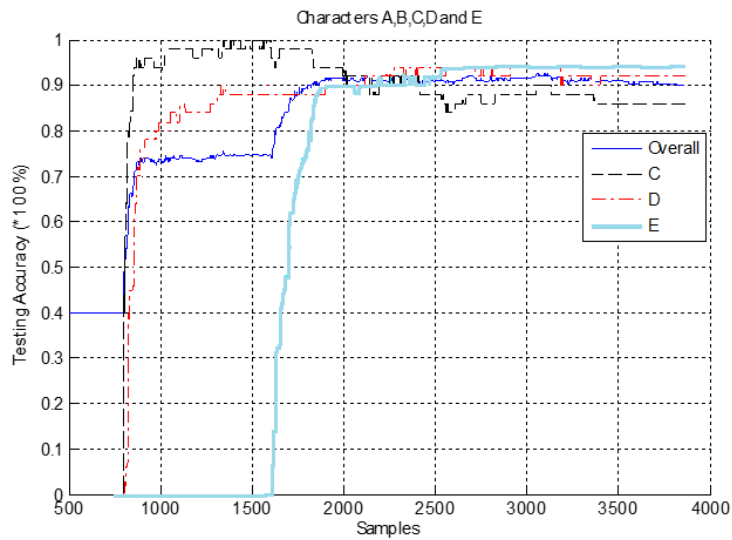

Fig. 11. Individual and Overall Testing Accuracy for Simultaneous New Classes

The proposed algorithm introduces new neurons in the output layer and recalibrates the network by itself to facilitate learning of new classes. Since only the output layer neurons are increased and the number of hidden layer neurons is the same, the learning of new classes that can be progressively learnt is limited by the number of classes that can be learnt by the given number of hidden layer neurons. Further, the proposed algorithm can be extended such that both the output neurons and hidden layer neurons are increased such that any number of new classes can be learnt progressively.

## 6.   CONCLUSIONS

In this paper, a novel learning technique of progressive learning for multi-class classification is developed. Progressive learning enables the network to learn multiple new classes dynamically on the run. The new classes can be learnt in both sequential and simultaneous manner. Hence this technique is much suited for applications where the number of classes to be learned is unknown. Progressive learning enables the network to recalibrate and adapt when encountered with a new class of data. The proposed progressive learning technique will perform effectively in applications such as cognitive robotics where the system is trained by real time experienced based data.

**ACKNOWLEDGMENT**

The first author would like to thank Nanyang Technological University, Singapore for the NTU Research Student Scholarship.

## REFERENCES

[1] D. E. Rumelhart, G. E. Hinton, R. J. Williams, "Learning Representations by Back-propagating Errors", *Nature*, vol. 323, no. 9, pp. 533-536, 1986.

[2] M. T. Hagan, M. B. Menhaj, "Training Feedforward Networks with the Marquardt Algorithm", *IEEE Trans. on Neural Networks*, vol. 5, no. 6, pp. 989-993, 1994.

[3] B. M. Wilamowski, H. Yu, "Neural Network Learning without Backpropagation", *IEEE Trans. on Neural Networks*, vol. 21. No. 11, pp. 1793-1803, 2010.

[4] S. Chen, C. Cowan, P. Grant, "Orthogonal Least Squares Learning Algorithm for Radial Basis Function Networks", *IEEE Trans. on Neural Networks*, vol. 2, no. 2, pp. 302-309, 1991.

[5] K. Li, J. X. Peng, G. W. Irwin, "A Fast Nonlinear Model Identification Method", *IEEE Trans. on Automatic Control*, vol. 50, no. 8, pp. 1211-1216, 2005.

[6] J. Branke, "Evolutionary Algorithms for Neural Network Design and Training", *Proc. Of first Nordic workshop on genetic algorithms and its applications*, 1995.

[7] X. Yao, "A Review of Evolutionary Artificial Neural Networks", *International Journal of Intelligent Systems*, vol. 8, no. 4, pp. 539-567, 1993.

[8] B. Li, J. Wang, Y. Li, Y. Song, "An Improved On-Line Sequential Learning Algorithm for Extreme Learning Machine", *Advances in Neural Networks – ISNN 2007*, vol. 4491, pp. 1087-1093, 2007.

[9] G. B. Huang, Q. Y. Zhu, and C. K. Siew, "Extreme Learning Machine: Theory and applications", *Neurocomputing*, vol. 70, pp. 489-501, 2006.

[10] D. E. Rumelhart, G. E. Hinton, R. J. Williams, "Learning Internal Representations by Error Propagation", No. ICS-8506, California University San Diego La Jolla Inst. for Cognitive Science, 1985.

[11] S. Ferrari, R. F. Stengel, "Smooth Function Approximation Using Neural Networks", *IEEE Transactions on Neural Networks*, vol. 16, pp. 24-38, 2005.

[12] G. B. Huang, Y. Q. Chen, H. A. Babri, "Classification Ability of Single Hidden Layer Feedforward Neural Networks", *IEEE Transactions on Neural Networks*, vol. 11, pp. 799-801, 2000.

[13] G. B. Huang, D. H. Wang, Y. Lan, "Extreme Learning Machines: A Survey", *International Journal of Machine Learning and Cybernetics*, vol. 2, pp. 107-122, 2011.

[14] N. Wang, M. J. Er, M. Han, "Generalized Single-hidden Layer Feedforward Networks for Regression Problems", *IEEE Transactions on Neural Networks and Learning Systems*, vol. 26, no. 6, pp. 1161-1176, 2015.

[15] N. Y. Liang, G. B. Huang, P. Saratchandran, N. Sundararajan, "A Fast and Accurate Online Sequential Learning Algorithm for Feedforward Networks", *IEEE Transactions on Neural Networks*, vol. 17, pp. 1411-1423, 2006.

[16] G. B. Huang, Q. Y. Zhu, C. K. Siew, "Extreme Learning Machine: A New Learning Scheme of Feedforward Neural Networks", *Proceedings of International Joint Conference on Neural Networks*, vol. 2, pp. 985-990, 2004.

[17] G. B. Huang, L. Chen, C. K. Siew, "Universal Approximation Using Incremental Constructive Feedforward Networks with Random Hidden Nodes", *IEEE Transactions on Neural Networks*, vol. 17, pp. 879-892, 2006.

[18] G. B. Huang, L. Chen, "Convex Incremental Extreme Learning Machine", *Neurocomputing*, vol. 70, no. 16, pp. 3056-3062, 2007.

[19] G. B. Huang, L. Chen, "Enhanced Random Search based Incremental Extreme Learning Machine", *Neurocomputing*, vol. 71, no. 16, pp. 3460-3468, 2008.

[20] Y. Wang, F. Cao, Y. Yuan, "A Study on Effectiveness of Extreme Learning Machine", *Neurocomputing*, vol. 74, pp. 2483-2490, 2011.

[21] G. B. Huang, H. Zhou, X. Ding, R. Zhang, "Extreme Learning Machine for Regression and Multiclass Classification", *IEEE Trans. on Systems, Man and Cybernetics, Part B: Cybernetics*, vol. 42, no. 2, pp. 513-529, 2012.

[22] Q. Y. Zhu, A. K. Qin, P. N. Suganthan, G. B. Huang, "Evolutionary Extreme Learning Machine", *Pattern Recognition*, vol. 38, pp. 1759-1763, 2005.

[23] M. B. Li, G. B. Huang, P. Saratchandran, N. Sundararajan, "Fully Complex Extreme Learning Machine", *Neurocomputing*, vol. 68, pp. 306-314, 2005.

[24] N. Wang, M. J. Er, M. Han, "Parsimonious Extreme Learning Machine Using Recursive Orthogonal Least Squares", *IEEE Transactions on Neural Networks and Learning Systems*, vol. 25, no. 10, pp. 1828-1841, 2014.

[25] G. B. Huang, P. Saratchandran, N. Sundararajan, "An Efficient Sequential Learning Algorithm for Growing and Pruning RBF (GAP-RBF) Networks", *IEEE Transactions on Systems, Man, and Cybernetics, Part B*: vol. 34, pp. 2284-2292, 2004.

[26] G. B. Huang, P. Saratchandran, N. Sundararajan, "A Generalized Growing and Pruning RBF (GGAP-RBF) Neural Network for Function Approximation", *IEEE Transactions on Neural Networks*, vol. 16, pp. 57-67, 2005.

[27] N. Wang, J. C. Sun, M. J. Er, Y. C. Liu, "Hybrid Recursive Least Squares Algorithm for Online Sequential Identification Using Data Chunks", *Neurocomputing*, vol. 174, pp. 651-660, 2016.

[28] N. Wang, M. Han, N. Dong, M. J. Er, "Constructive Multi-output Extreme Learning Machine with Application to Large Tanker Motion Dynamics Identification", *Neurocomputing*, vol. 128, pp. 59-72, 2014.

[29] H. J. Rong, G. B. Huang, N. Sundararajan, P. Saratchandran, "Online Sequential Fuzzy Extreme Learning Machine for Function Approximation and Classification Problems", *IEEE Transactions on Systems, Man, and Cybernetics, Part B:* vol. 39, pp. 1067-1072, 2009.

[30] G. B. Huang, N. Y. Liang, H. J. Rong, P. Saratchandran, N. Sundararajan, "On-Line Sequential Extreme Learning Machine", *Computational Intelligence,* vol. 2005, pp. 232-237, 2005.

[31] G. Huang, G. B. Huang, S. Song, K. You, "Trends in Extreme Learning Machines: A Review", *Neural Networks*, vol. 61, pp. 32-48, 2015.

[32] M. R. Daliri, "A Hybrid Automatic System for the Diagnosis of Lung Cancer Based on Genetic Algorithm and Fuzzy Extreme Learning Machines", *Journal of Medical Systems*, vol. 36, no. 2, pp. 1001-1005, 2012.

[33] W. B. Zhang, H. B. Ji, "Fuzzy Extreme Learning Machine for Classification", *Electronics Letters*, vol. 49, no. 7, pp. 448-449, 2013.

[34] E. Avci, R. Coteli, "A New Automatic Target Recognition System based on Wavelet Extreme Learning Machine", *Expert Systems with Applications*, vol. 39, no. 16, pp. 12340-12348, 2012.

[35] V. Malathi, N. S. Marimuthu, S. Baskar, K. Ramar, "Application of Extreme Learning Machine for Series Compensated Transmission Line Protection", *Engineering Applications of Artificial Intelligence*, vol. 24, no. 5, pp. 880-887, 2011.

[36] B. Freney, M. Verleysen, "Parameter-insensitive Kernel in Extreme Learning for Non-linear Support Vector Regression", *Neurocomputing*, vol. 74, no. 16, pp. 2526-2531, 2011.

[37] W. W. Zong, G. B. Huang, Y. Q. Chen, "Weighted Extreme learning Machine for Imbalance Learning", *Neurocomputing*, vol. 101, pp. 229-242, 2013.

[38] Z. H. Man, K. Lee, D. H. Wang, Z. W. Cao, C. Y. Miao, "A New Robust Training Algorithm for a Class of Single Hidden Layer feedforward neural networks", *Neurocomputing*, vol. 74, no. 16, pp. 2491-2501, 2011.

[39] J. W. Cao, Z. P. Lin, G. B. Huang, N. Liu, "Voting based Extreme Learning Machine", *Information Sciences*, vol 185, no. 1, pp. 66-77, 2012.

[40] Z. H. You, Y. K. Lei, L. Zhu, J. F. Xia, B. Wang, "Prediction of Protein-protein Interactions from Amino Acid Sequences with Ensemble Extreme Learning Machines and Principal Component Analysis", *BMC Bioinformatics*, 14, 2013.

[41] J. H. Zhai, H. Y. Xu, X. Z. Wang, "Dynamic Ensemble Extreme Learning Machine based on Sample Entropy", *Soft Computing*, vol. 16, no. 9, pp. 1493-1502, 2012.

[42] Jarvis, Peter. "Towards a Comprehensive Theory of Human Learning*"*, Vol. 1, *Psychology Press*, 2006.

[43] Z. Zhao, Z. Chen, Y. Chen, S. Wang, H. Wang, "A Class Incremental Extreme learning machine for activity recognition", *Cognitive Computation*, vol. 6 no. 3, pp. 423-431, 2014.